\documentclass[letterpaper, 10 pt, conference]{ieeeconf}  

\IEEEoverridecommandlockouts                              

\usepackage{graphics} 
\usepackage{epsfig} 
\usepackage{mathptmx} 
\usepackage{times} 
\usepackage{amsmath} 
\usepackage{amssymb}  
\usepackage{algorithm}
\usepackage{algpseudocode}
\usepackage{caption}
\title{\LARGE \bf
 RRT and Velocity Obstacles-based motion planning for Unmanned Aircraft Systems Traffic Management (UTM)}

\author{\authorblockN{Himanshu, Jinraj V Pushpangathan, and Harikumar Kandath} \\
\authorblockA{\textit{International Institute of Information Technology Hyderabad, India} \\
\textit{himanshukmr234@gmail.com; jinrajaero@gmail.com; harikumar.k@iiit.ac.in}}}

\begin{document}

\maketitle
\thispagestyle{empty}
\pagestyle{empty}

\begin{abstract}

In this  paper, an algorithm for Unmanned Aircraft Systems Traffic Management (UTM) for a finite number of unmanned aerial vehicles (UAVs) is proposed. This algorithm is developed by combining the Rapidly-Exploring Random Trees (RRT) and Velocity Obstacle (VO) algorithms and is referred to as the RRT-VO UTM algorithm. Here, the RRT algorithm works offline to generate obstacle-free waypoints in a given environment with known static obstacles. The VO algorithm, on the other hand, operates online to avoid collisions with other UAVS and known static obstacles. The boundary of the static obstacles are approximated by small circles to facilitate the formulation of VO algorithm. The proposed algorithm's performance is evaluated using numerical simulation and then compared to the well-known artificial potential field (APF)  algorithm for collision avoidance. The advantages of the proposed method are  clearly shown in terms of lower path length and collision avoidance capabilities for a challenging scenario.

\end{abstract}

\section{INTRODUCTION}

Over the last decade, unmanned aerial vehicles (UAVs) have been widely used in both the military (intelligence, surveillance and reconnaissance missions) \cite{jj} and civilian (agriculture, mapping and surveying, entertainment, deliveries, disaster relief missions etc.) sectors \cite{hari1}. This sudden increase in UAVs poses a serious risk to civilians on the ground, as well as piloted aircraft and other UAVs operating in the same airspace. The rise in the number of UAVs in a given airspace necessitates a system referred to as the Unmanned Aircraft Systems Traffic Management (UTM) system that monitors and
manages these UAVs, especially at low altitudes. The UTM’s objective is to provide a safe, efficient, and secure airspace management system for the UAVs in the same airspace.\par

The primary responsibility of UTM is to ensure that UAVs keep a safe distance from other aircraft and obstacles. To accomplish this, several methods are developed. The graph-based methods are used in \cite{bae2018new} to plan the flight paths of multiple UAVs and maintain a minimum distance between them. In \cite{ong2015short}, the problem of collision avoidance is solved by decomposing a large Markov decision process (MDP) into small MDPs to find a solution offline and then combined online iteratively to produce a locally optimal solution. The A* algorithm is utilized in \cite{tan2019evolutionary} to generate flight paths for multiple UAVs in urban airspace. Thereafter, an evolutionary algorithm is employed to schedule the flights to avoid conflicts.  One of the most commonly used algorithms in UTM is the Artificial Potential Field (APF). The dynamic artificial potential field was used for collision avoidance in \cite{du2019realDynamicAPF}. In \cite{HAPTICAPF}, authors investigate the use of modified APF along with a haptic-based tele-operation. Authors of \cite{conflictAPF} use the APF and priority allocation rules for conflict resolution in UAVs. Trajectory prediction method and APF was used for maintaining a safe distance between UAVs and non cooperative obstacles in \cite{du2019automatic}. A computationally intensive collision avoidance algorithm using predictive control is proposed in \cite{ishaan} to avoid collision among multiple UAVs in the context  of UTM. 
\par

The main contribution of this paper is the new UTM algorithm, termed the RRT-VO UTM algorithm, for a finite number of UAVs operating in a 2D environment with known 2D static obstacles. This  algorithm combines elements of an offline path planning algorithm and an online collision avoidance algorithm to enable efficient and safe coordination of multiple UAVs. Here, the offline path planning is accomplished using the Rapidly Exploring Random Tree (RRT)  algorithm  \cite{lavalle1998rapidlyRRT} and Velocity Obstacle (VO) algorithm \cite{fiorini1993motionVO}  provides collision avoidance. The RRT is chosen among the existing offline path planning algorithms due to its efficient exploratory ability. Similarly, the Velocity Obstacle (VO) is used for collision avoidance due to its ease of implementation and ability to handle multiple UAVs, making it well-suited for the UTM. To avoid collisions, the VO algorithm in the RRT-VO UTM algorithm searches for a suitable velocity vector for each UAV. In the proposed algorithm, the boundaries of 2D known static obstacles are represented using overlapping circles in order to apply the VO algorithm against the 2D known static obstacles.  The proposed algorithm's effectiveness is demonstrated by generating feasible paths for five quadcopters operating in the same 2D environment with known static obstacles. Also, the collision avoidance among UAVs and UAV to static obstacles is demonstrated for navigation through a challenging environment. A comparison is provided with the well known artificial potential filed (APF) method. The APF method fail to  avoid collision, whereas using the proposed method, all the UAVs were able to navigate successfully avoiding collisions.
\par
In Velocity Obstacle technique there are several ways to find a safe velocity to prevent collisions. In \cite{fioriniVO}, the authors describe two main methods: a global search and a heuristic search. The global search method is a comprehensive approach that finds the safe velocity by building a tree of possible maneuvers over time intervals. The heuristic search, on the other hand, is a more targeted approach that only searches for safe velocities at one time step using certain rules, or heuristics, to choose the best velocity from the options available. These heuristics include "To Goal", which prioritizes reaching the target with the highest velocity along the direct path; "Maximum Velocity", which chooses the highest avoidance velocity within a certain angle from the line to the goal; and "Structure", which selects the velocity that best avoids obstacles based on the level of risk they pose. \par

Our proposed RRT-VO UTM algorithm uses a hybrid approach that blends the "Structure" and "To Goal" heuristics. This means it tries to reach the goal as quickly as possible while taking into account the presence of obstacles. If an obstacle is detected, the algorithm will choose a velocity that not only avoids the obstacle but is also similar to the original velocity that was heading toward the goal. This hybrid approach considers both the direction and speed of the new velocity, allowing the unmanned aerial vehicle (UAV) to slow down or stop if necessary.

The rest of this paper is organized as follows: Section II gives the preliminaries and problem formulation. Section \ref{MA} describes the development of the RRT-VO UTM algorithm using the RRT and VO algorithms. The simulation results are discussed in Section \ref{SR}. The key findings of this paper and future research directions are summarized in Section \ref{CL}. 

\section{Preliminaries and Problem formulation}
\begin{figure}[h]
\begin{center}
\includegraphics[width= 8cm]{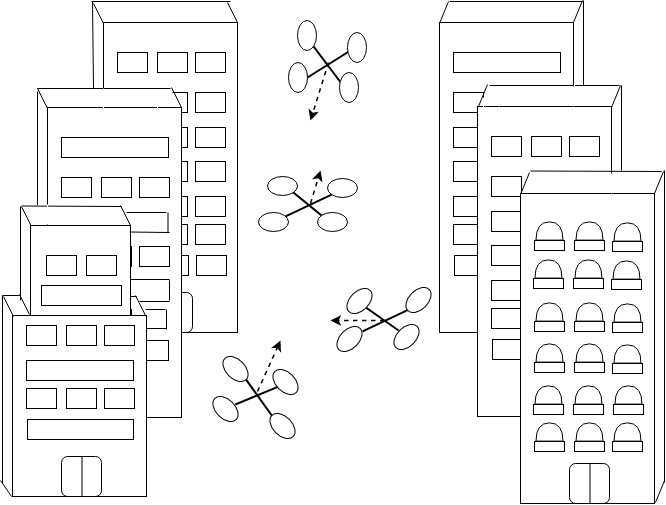}    
\caption{UAVs trying to navigate the airspace with other UAVs and buildings.} 
\label{fig:utm}
\end{center}
\end{figure}

The increase in the volume of UAVs used for urban applications like package delivery, air ambulance etc. (as shown in Figure \ref{fig:utm}) demands low-altitude flight. Efficient low-altitude flight in the presence of a large number of buildings and multiple UAVs  requires algorithms that can generate optimal offline paths and also real-time collision avoidance. Static obstacles like buildings are approximated by rectangles (as illustrated in Figure \ref{fig:RRT}) for offline path generation. \\

The UAVs considered are here of quadrotor type with first order velocity control implemented in the outer loop as given in equations (\ref{quad1}) and (\ref{quad2}) for X-axis and Y-axis respectively.
\begin{equation}
  \dot{X}=V_x  
  \label{quad1}
\end{equation}
\begin{equation}
  \dot{Y}=V_y  
  \label{quad2}
\end{equation}
Here $[X,\,Y]$ are the 2-dimensional coordinates of the UAV with respect to the inertial frame and the control input is the velocity given by $\textbf{V}=[V_x,\,V_y]^T$. The following assumptions are made in this paper. \\

\begin{enumerate}
    \item The obstacles in the environment are static and their positions are known.
    \item The known static obstacles are approximated by rectangles. 
    \item The position and velocity of each UAV are broadcasted to all other UAVs.
    \item Furthermore, the UAVs are approximated by a circle with a given radius from their centre of gravity location.
    \item The starting and end waypoint location for each UAV is known and is given as input to the offline planner.
    \item The UAVs fly at a constant altitude from the ground. Thus, the motion planning is done in 2D space.

\end{enumerate}
The first objective here is to generate a safe offline path for UAVs to traverse from starting location to the destination location avoiding static obstacles. The second  objective is to avoid UAV-to-UAV collision as well as UAV-to-static obstacle collision by generating a  safe trajectory online, with a low computational complexity.
\section{ RRT-VO  UTM Algorithm}\label{MA} 

In this section, a new UTM algorithm called the RRT-VO UTM algorithm based on the RRT and VO algorithms is presented. This algorithm's goal is to generate feasible paths for N numbers of UAVs that operate in the same environment (airspace) with known static obstacles. The feasible path of a UAV is the one that has  starting and goal positions and does not collide with known static obstacles or other UAVs. The RRT-VO UTM algorithm generally has an offline path planning module and a collision avoidance module. Given the environment with known static obstacles, the RRT algorithm acts as an offline path planner that determines a path that avoids collision with known static obstacles for UAVs before their take off and starting their mission. The VO algorithm, on the other hand, is used in real-time to avoid collisions between UAVs and static obstacles
in the environment. Next, the  RRT algorithm is briefly described.
\subsection{Rapidly-Exploring Random Tree Algorithm}
\begin{figure}[h]
\begin{center}
\includegraphics[width= 8cm]{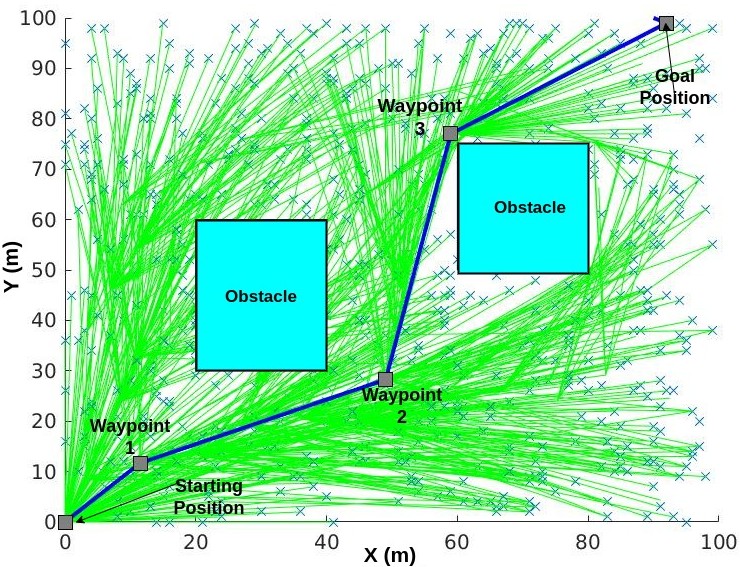}    
\caption{Offline path generated by RRT algorithm for navigation of UAV from starting to goal position.} 
\label{fig:RRT}
\end{center}
\end{figure}

Rapidly-Exploring Random Tree (RRT) is a sampling-based motion planning algorithm that is well-suited for high-dimensional complex configuration spaces (environment) with a large number of obstacles or other constraints. The main idea behind RRT is to construct a tree-like structure in the configuration
space by incrementally adding vertices that represent the possible states of the system. In each iteration, a new random configuration is chosen, and the nearest vertex in the tree is found. The algorithm then attempts to move towards the random configuration from the nearest vertex, creating a new vertex in the tree if the new configuration is in free space and not in collision. The process is
repeated until a path between the start and goal configurations is found. Figure  \ref{fig:RRT} shows the path generated by the RRT algorithm. In this figure, the tree-like structure is represented with green lines, and the small blue crosses represent the random configuration. The red squares represent the waypoints generated by the RRT algorithm.

\subsection{Velocity Obstacle Algorithm}

The Velocity Obstacle algorithm considers the relative motion of other UAVs in the environment and determines a set of safe velocities for the UAVs to avoid collisions. This algorithm creates a virtual "cone" around each UAV based on its current velocity and position, as well as the velocities and positions of other UAVs/obstacles in the environment. The algorithm then finds the set of safe velocities for the UAV to avoid collisions with all of the other UAVs and static obstacles by finding the intersection of the feasible velocities for the UAV. This intersection is known as the "safe velocity set" and the UAV can choose its velocity from this set to avoid collision.\par
\begin{figure}[h!]
\begin{center}
\includegraphics[width=8cm]{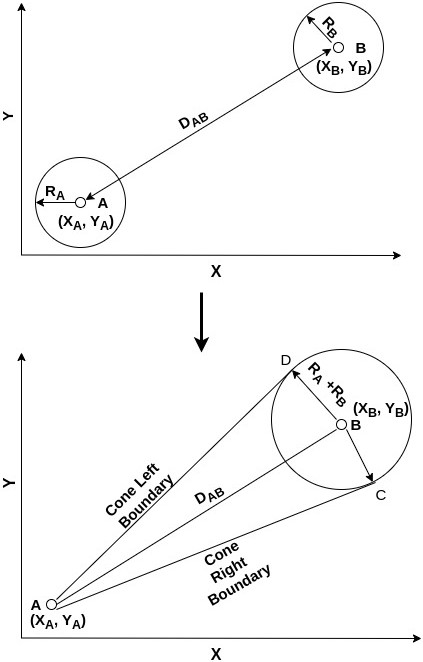}    
\caption{Formation of Collision cone between the UAV A and UAV B (or Obstacle B).} 
\label{fig:cone}
\end{center}
\end{figure}

Figure \ref{fig:cone} shows the collision cone formed between UAV A and UAV B. The collision cone, which is used to determine potential collisions between a UAV (referred to as UAV A) and other UAVs (UAV B) or obstacles (Obstacle B), is defined using two tangents on a circle from an external point. The radius of this circle is the sum of the radii of UAV A and UAV B (Obstacle B). Let $\mathbf{V_{A}}$ be  the velocity of UAV A with respect to (w.r.t)  a reference frame, $\mathbf{V_{B}}$ be the velocity of UAV B  w.r.t  the same reference frame, and $\mathbf{V_{AB}}$ be the relative velocity vector of UAV A w.r.t UAV B. If $\mathbf{V_{AB}}$ falls within the  collision cone, then a collision between UAV A and UAV B is guaranteed. To prevent this collision, the algorithm selects the $\mathbf{V_{AB}}$ that does not belong to this collision cone. Using this new $\mathbf{V_{AB}}$, the algorithm calculates new velocities ($\mathbf{V_{A}}$) for UAV A as potential solutions. 
To compute cone, let $\angle ABX$, $\angle DAB$, and $\angle CAB$ denote the angle between the line AB and the X-axis, the angle between the lines AD and AB, and the angle between the lines CA and AB, respectively. Moreover, the position of UAV A in 2D space is represented by the coordinates $(X_{A}, Y_{A})$, and the position of UAV B is represented by the coordinates $(X_{B}, Y_{B})$. Also, the radius of UAV A and B is represented by $R_{A}$ and $R_{B}$,  respectively. Moreover,  $\angle C_{left}$ symbolizes  the angle of left  cone boundaries and $\angle C_{right}$  exemplifies the angle of right cone boundaries. $D_{AB}$ represents distance between A and B. Now, the collision cone is computed as per equations given below.    
\begin{equation}
\begin{aligned}
\angle ABX =arctan\bigg(\frac{Y_{B}-Y_{A}}{X_{B}-X_{A}}\bigg)
\label{eq:1}
\end{aligned}
\end{equation}

\begin{equation}
\begin{aligned}
\angle DAB = \angle CAB  = arcsin\bigg(\frac{R_{A}+R_{B}}{D_{AB}}\bigg)
\label{eq:2}
\end{aligned}
\end{equation}

\begin{equation}
\begin{aligned}
\angle C_{left} = \angle ABX + \angle DAB
\label{eq:3}
\end{aligned}
\end{equation}

\begin{equation}
\begin{aligned}
\angle C_{right} = \angle ABX - \angle CAB
\label{eq:4}
\end{aligned}
\end{equation}

Here Eqn. \ref{eq:1} is for calculating $\angle ABX$, Eqn. \ref{eq:2} calculates $\angle DAB$ and $\angle CAB$. Equation \ref{eq:3} and \ref{eq:4} is for calculating the angle for left $\angle C_{left}$ and right $\angle C_{right}$ cone boundaries.
\subsection{RRT-VO UTM Algorithm}
\begin{figure}[h!]
\begin{center}
\includegraphics[width= 7cm]{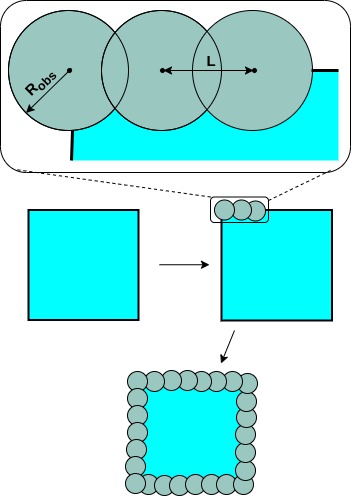}    
\caption{Representation of a rectangular obstacle using circular obstacles to apply Velocity obstacle algorithm for collision avoidance with known static obstacles.} 
\label{fig:Obstacle}
\end{center}
\end{figure}

\noindent The RRT-VO UTM algorithm  is presented in this section. To explain  the RRT-VO UTM algorithm, first assume that the starting and goal positions of all the UAVs are known. Following that, the environment is formed  by placing the obstacles in a rectangular shape. The RRT algorithm is then used offline to generate obstacle-free (collision-free with known static obstacles) waypoints from the starting and goal positions for all the UAVs in this environment. Collisions between known static obstacles and  UAVs can be avoided if the UAVs follow these waypoints. However, collisions between the UAVs may happen. Also, if UAVs attempt to avoid these collisions near static obstacles, they may collide with known static obstacles in the environment. Hence, the UTM requires collision-avoidance capabilities that can also avoid these 2D static obstacles. The velocity obstacle (VO) method is used in the RRT-VO UTM algorithm for  collision avoidance. To provide the desired collision-avoidance capabilities to the RRT-VO UTM algorithm using the velocity obstacle method,  rectangular static obstacles are represented using circular obstacles of radius $R_{obs}$ with $L$ distance apart, as shown in Fig.
\ref{fig:Obstacle}. This figure depicts the boundary of a known static obstacle as an overlapping circle. The RRT-VO UTM algorithm can compute these circles as the static obstacle's boundary and circle radius are known in advance. Therefore, using the circular representations of the known static obstacle's boundaries, the VO algorithm is capable of avoiding collisions between UAVs even when they are close to known static obstacles. Likewise, each UAVs are represented as circular obstacles of radius $R_A+R_B$ as shown in Fig. \ref{fig:cone} for  collision avoidance between UAVs. \par
After converting the boundaries of rectangular obstacles into overlapping circular obstacles, the new UTM algorithm assigns the first waypoint to each UAV and calculates the velocity for tracking the assigned waypoint. Subsequently, the UTM algorithm computes the relative distance between each UAV using the position information of all the UAVs. When the relative distance of  one UAV, say UAV A, from some other UAV,  say UAV B, is less than a specific value ($Dist_{uav}$), then the relative velocity $\mathbf{V_{AB}}$  is computed and checked if this velocity is inside the collision cone angles [$\angle C_{left}$, $\angle C_{right}$]. If it fall within the cone, then the UTM algorithm  search for new relative velocity, $\mathbf{V_{AB}'}$. The UTM algorithm has a search loop to identify appropriate angle (denoted by $\theta$)  ranging from  of $0$ to $2\pi$ radians with a step size of $0.2$ radians for  the $\mathbf{V_{AB}'}$. Within this loop, another search loop is implemented to determine an appropriate magnitude of new relative velocity (represented by $M$),  $\mathbf{V_{AB}'}$,  ranging from $0$ to $\|\mathbf{V_{AB}}\|_{2}$ with step size of $0.2$. Now, for each $\theta$ and $M$,   $V_{AB}'$ constructed as per Eqn. \ref{eq:5}.

\begin{equation}
\begin{aligned}
\mathbf{V_{AB}'} = [M * cos(\theta), M * sin(\theta)] 
\label{eq:5}
\end{aligned}
\end{equation}
If $\mathbf{V_{AB}'}$ falls outside of the cone, then it is considered as the  potential solution and the new velocity for UAV A, $\mathbf{V_{A}' \in \mathbb{R}^{2}}$, is calculated with equation mentioned in Eqn. \ref{eq:6}.  
\begin{equation}
\begin{aligned}
\mathbf{V_{A}'} = [M * cos(\theta) + {\|\mathbf{V_{B}}\|_{2}} * cos(\angle \mathbf{V_{B}}),\\ M * sin(\theta) + {\|\mathbf{V_{B}}\|_{2}} * sin(\angle \mathbf{V_{B}})]
\label{eq:6}
\end{aligned}
\end{equation}
The computed values of feasible velocities are then stored in an array $fV_{A}$. Similarly,   the collision cones are constructed for all other UAVs or obstacles.However,  no further search for new $\mathbf{V_{AB}}$ is performed. Instead, each entry in $fV_{A}$ is evaluated, and the corresponding relative velocity $\mathbf{V_{AB}'}$ is calculated. If this $\mathbf{V_{AB}'}$ falls within the cone, the corresponding $\mathbf{V_{A}'}$ is removed from $fV_{A}$. After this process has been completed for all other UAVs and obstacles, the velocity $\mathbf{V_{A}'}$ is selected from $fV_{A}$ based on its similarity to original velocity $\mathbf{V_{A}}$, as determined by the Euclidean norm. Then, the position of the UAV A is updated by using the selected $\mathbf{V_{A}'}$ as given in Equation \ref{eq:7} and \ref{eq:8}.
\begin{equation}
\begin{aligned}
X_{A} = X_{A} + dt *VX_{A}' 
\label{eq:7}
\end{aligned}
\end{equation}

\begin{equation}
\begin{aligned}
Y_{A} = Y_{A} + dt *VY_{A}'
\label{eq:8}
\end{aligned}
\end{equation}

In Eqn. \ref{eq:7} \& \ref{eq:8}, $VX_{A}'$ and $VY_{A}'$ are the x and y components of $\mathbf{V_{A}'}$ and $dt$ is the time-step used for integrating the velocities. In the RRT-VO algorithm, the aforementioned process of determining a new velocity to avoid collision is repeated for each UAV in the airspace at each time-step. The pseudocode of the RRT-VO algorithm is given in Algorithm \ref{alg:RRT-VO}.
\begin{algorithm}[h!]
\caption{RRT-VO UTM Algorithm}
\label{alg:RRT-VO}
\begin{algorithmic}[1]
\State \textbf{Input:} UAV start, goal position $(X, Y)$ and initial \par velocity $(VX, VY)$ and obs. pos., width, height
\State Calculate waypoints for each UAV using RRT
\State Generate circular obstacles from rectangular obstacles
\State Assign initial waypoints to each UAV
\While{goals not reached}
\For{each UAV A}
\If{distance between A, waypoint$<Dist_{wp}$}
\State Assign next waypoint $(X_{Awp}, Y_{Awp})$ to A
\EndIf
\EndFor
\For{each UAV A}
\State Calculate velocity \par \hspace{0.3cm} $\mathbf{V_{A}}  = [kp*(X_{Awp} - X_{A}), kp*(Y_{Awp} - Y_{A})]$
\State Initialize feasible velocity set $fV_{A}$
\For{each remaining UAV B}
\If{distance between A, B $<Dist_{uav}$}
\State Calculate relative velocity \par \hspace{1.3cm} $\mathbf{V_{AB}} = [VX_{A} - VX_{B}, VY_{A} - VY_{B}]$
\State Calculate angle bound for collision cone \par \hspace{1.3cm} from Eqn. \ref{eq:3} and \ref{eq:4}
\If{$\mathbf{V_{AB}}$ lies in cone and $fV_{A}$ is empty}
\State Search for $\mathbf{V_{AB}'}$ outside cone by Eqn.\ref{eq:5}
\State Calculate $\mathbf{V_{A}'}$ from $\mathbf{V_{AB}'}$ using Eqn. \ref{eq:6} \par \hspace{1.8cm} and store in $fV_{A}$
\EndIf
\If{$\mathbf{V_{AB}}$ lies in cone \& $fV_{A}$ not empty}
\For{each $\mathbf{V_{A}'}$ in $fV_{A}$}
\State Calculate $\mathbf{V_{AB}'}$ from $\mathbf{V_{A}'}$
\If{$\mathbf{V_{AB}'}$ lies in cone}
\State Remove $\mathbf{V_{A}'}$ from $fV_{A}$
\EndIf
\EndFor
\EndIf
\EndIf
\EndFor
\For{each obstacle B}
\If{distance between A, B$<Dist_{obs}$}
\State \textbf{Repeat} steps 16-29
\EndIf
\EndFor
\State Select $\mathbf{V_{A}'}$ from $fV_{A}$ similar to original $\mathbf{V_{A}}$  
\State Update UAV A's position using $\mathbf{V_{A}'}$ from \par \hspace{0.3cm} Eqn. 7 and 8
\EndFor
\EndWhile
\end{algorithmic}
\end{algorithm}

\section{SIMULATION RESULTS}\label{SR}
In this section, the effectiveness of the RRT-VO UTM algorithm is evaluated for multiple UAVs operating in the same environment with a dimension of 400~m $\times$ 400~m. The environment is made up of multiple rectangular obstacles that are converted into overlapping circular obstacles. The performance of the proposed algorithms is assessed in terms of safety \& collision-free trajectory generation and path length for five and seven quadcopters in the given environment with known static obstacles. For comparison with the proposed algorithm, the Artificial Potential Field (APF)-based UTM algorithm is used to generate paths for these UAVs operating in the same environment. APF works on the basis of attraction force to reach goal/waypoint and repulsion force to avoid collisions with other UAVs and obstacles. Suppose $(X_{A}, Y_{A})$ is the position of UAV A, $(X_{Awp}, Y_{Awp})$ as the position of assigned waypoint for UAV A and $(X_{B}, Y_{B})$ is the position of other UAV or obstacle B. The equations for calculating attractive and repulsive force are mentioned in Eqn. \ref{eq:10} \& \ref{eq:12}.

\begin{equation}
\begin{aligned}
Dir_{att} = [X_{Awp} - X_{A}, Y_{Awp} - Y_{A}]
\label{eq:9}
\end{aligned}
\end{equation}

\begin{equation}
\begin{aligned}
F_{att} = k_{att} * \frac{Dir_{att}}{\|Dir_{att}\|}
\label{eq:10}
\end{aligned}
\end{equation}

\begin{equation}
\begin{aligned}
Dir_{rep} = [X_{A} - X_{B}, Y_{A} - Y_{B}]
\label{eq:11}
\end{aligned}
\end{equation}

\begin{equation}
\begin{aligned}
F_{rep} = k_{rep} * \frac{Dir_{rep}}{\|Dir_{rep}\|}
\label{eq:12}
\end{aligned}
\end{equation}

Here, Eqn. \ref{eq:9} \& \ref{eq:11} calculates the direction vector ($Dir_{att}$  \& $Dir_{rep}$) for attraction  and repulsion force respectively.\par
The attraction and repulsion force are added together to calculate total force (shown in Eqn \ref{eq:13}) which is then used to update the position of UAV A as shown in Eq. \ref{eq:14} \& \ref{eq:15}. 

\begin{equation}
\begin{aligned}
TF = F_{att} + F_{rep}
\label{eq:13}
\end{aligned}
\end{equation}

\begin{equation}
\begin{aligned}
X_{A} = X_{A} + dt*TF_{X}
\label{eq:14}
\end{aligned}
\end{equation}

\begin{equation}
\begin{aligned}
Y_{A} = Y_{A} + dt*TF_{Y}
\label{eq:15}
\end{aligned}
\end{equation}
where $TF_{X}$ and $TF_{Y}$ are the x and y component of total force ($TF$). Position of each UAV is updated similarly using equations \ref{eq:9}-\ref{eq:15} to avoid obstacles and other UAVs. \par

The RRT-VO UTM algorithm (\textbf{Algorithm \ref{alg:RRT-VO}}) is implemented in Matlab and uses the parameters given in Table \ref{tb:parameter1} for the simulations. The APF-based UTM algorithm uses the same waypoints generated by the RRT algorithm. It also uses basic mechanism such as activation of collision avoidance when another UAV or obstacle is below certain distance ($Dist_{uav}$/$Dist_{obs}$) from current UAV and assigning a new waypoint when distance of UAV from previous waypoint is less than a certain value ($Dist_{wp}$). The parameters used in APF algorithm are mentioned in Table \ref{tb:parameter2}.

\begin{table}[h!]
\begin{center}
\caption{Parameters for simulating the RRT-VO UTM algorithm}\label{tb:parameter1}
\begin{tabular}{|c||c|}
\hline
Parameter & Value \\\hline
$k_{p}$ & 0.2\\
$dt$ & 0.1 \\
$R_{uav}$ & 12 \\
$R_{obs}$ & 12 \\
$L$ & 15 \\
$Dist_{wp}$ & 10 \\
$Dist_{uav}$ & 50 \\
$Dist_{obs}$ & 20 \\\hline

\end{tabular}
\end{center}
\end{table}

\begin{table}[h!]
\begin{center}
\caption{Parameters for simulating the RRT-APF UTM algorithm}\label{tb:parameter2}
\begin{tabular}{|c||c|}
\hline
Parameter & Value \\\hline
$k_{att}$ & 8\\
$k_{rep}$ & 15\\
$dt$ & 0.1 \\
$Dist_{wp}$ & 10 \\
$Dist_{uav}$ & 50 \\
$Dist_{obs}$ & 20 \\\hline

\end{tabular}
\end{center}
\end{table}

\begin{figure}[h!]
\begin{center}
\includegraphics[width=8cm]{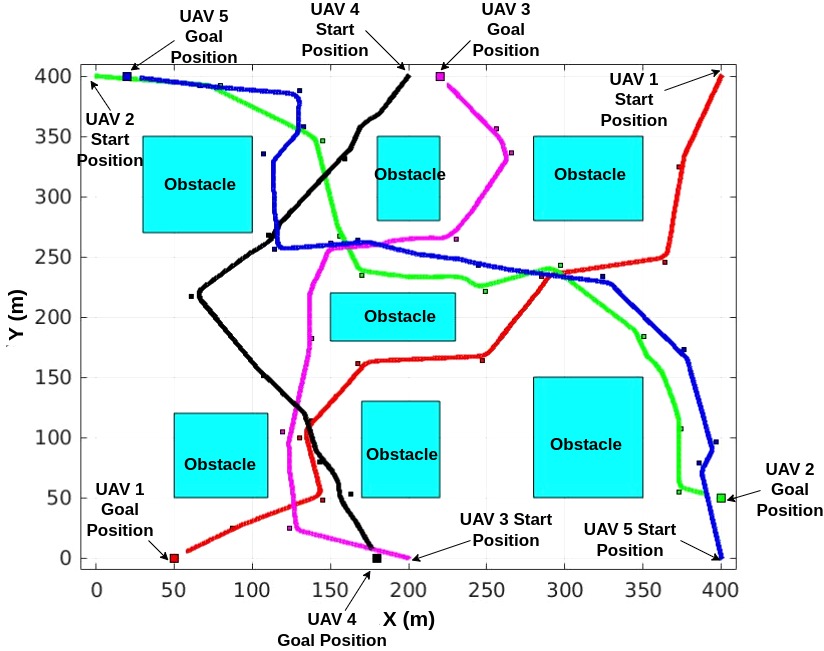}
\caption{Five UAVs navigating the environment using RRT-VO UTM algorithm.} 
\label{fig:5_UAV}
\end{center}
\end{figure}

\begin{figure}[h!]
\begin{center}
\includegraphics[width=8cm]{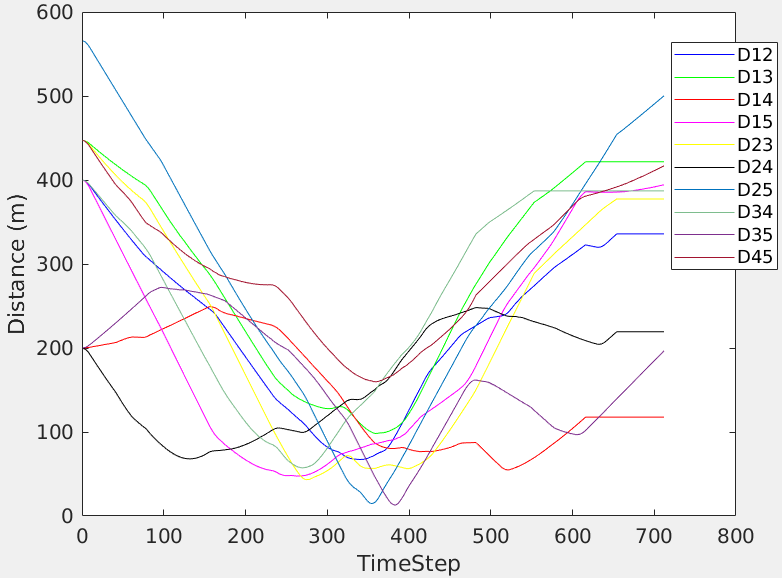}
\caption{Distance between the pair of UAVs while navigating the environment using RRT-VO UTM algorithm.} 
\label{fig:5_UAV_Distance}
\end{center}
\end{figure}

\begin{figure}[h!]
\begin{center}
\includegraphics[width=8cm]{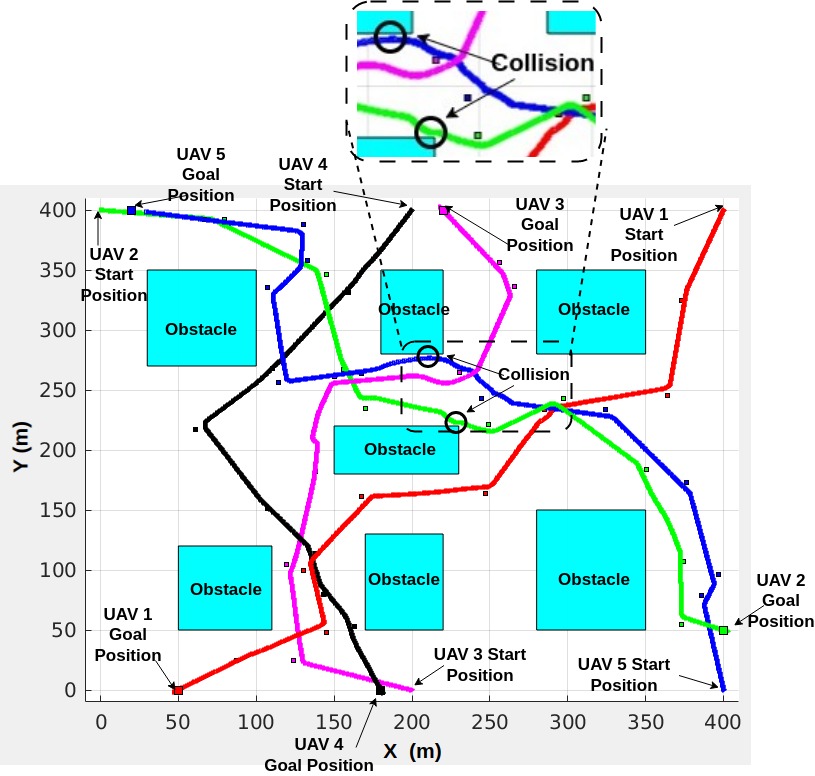}     
\caption{Five UAVs navigating the environment using RRT-APF UTM algorithm. UAV 2 and UAV 5 collides with edge of the static obstacle. Top figure is the magnified image of the area in which collision occurs.} 
\label{fig:5_UAVb}
\end{center}
\end{figure}

\begin{figure}[h!]
\begin{center}
\includegraphics[width=8cm]{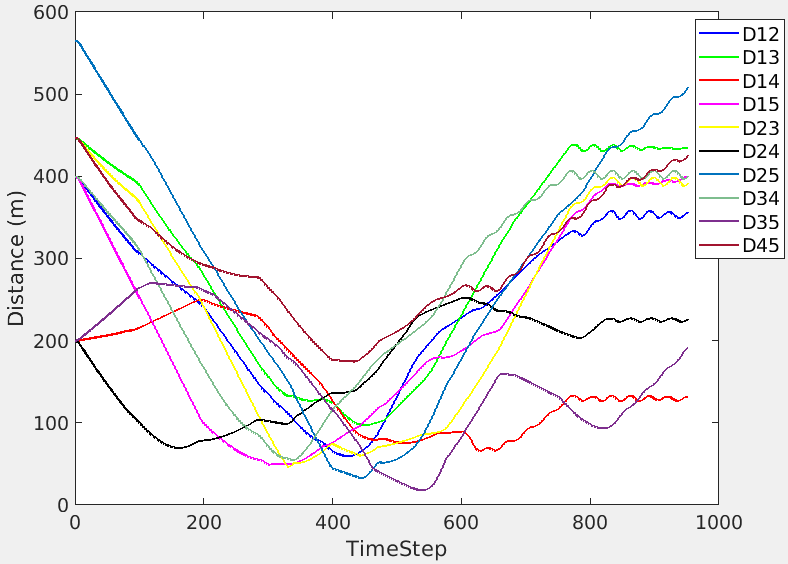}
\caption{Distance between the pair of UAVs while navigating the environment using RRT-APF UTM algorithm.} 
\label{fig:5_UAV_APF_Distance}
\end{center}
\end{figure}

The feasible paths generated by the RRT-VO UTM algorithm is shown in Fig. \ref{fig:5_UAV}. These paths are not colliding with any static obstacles. Also, the paths shown in Fig. \ref{fig:5_UAV} indicates that the proposed UTM algorithm is able to avoid collision with known static obstacles and other UAVs. This is substantiated with relative distance between UAVs shown in Fig. \ref{fig:5_UAV_Distance}. The relative distances are non-zero which suggest non collision between UAVs. For RRT-APF method it is clearly evident from non zero relative distance (shown in Fig. \ref{fig:5_UAV_APF_Distance}) that each UAV is able to avoid collision with other UAV. However, Fig. \ref{fig:5_UAVb} shows that UAV 2 and UAV 5 collided with obstacles (represented by circle) while trying to navigate through the environment.\par
Another point of comparison between the two algorithms is the path length. Table \ref{tb:PathLength} shows the path length of both the algorithms for the environment with five UAVs as shown in Fig. \ref{fig:5_UAV} and Fig. \ref{fig:5_UAVb}. The  "--" term in RRT-APF column for UAV 2 and 5 indicates the occurrence of a collision, thus ignoring the path length values for these cases. For each of the five UAVs, RRT-VO UTM algorithm is generating lower path length as compared to RRT-APF UTM algorithm. 
Similarly, Table \ref{tb:PathLength7} shows path length of seven UAVs navigating in the same environment. In this experiment, UAV 3, UAV 4 \& UAV 6 collided with the obstacle when using RRT-APF method. Thus, the path length of these UAVs is mentioned as "--". On the other hand, the UAVs using RRT-VO UTM algorithm have shorter path lengths without any collisions when compared with RRT-APF for seven UAVs.\par

In this paper, only quadcopters are used for simulations. However, it is important to note that the RRT-VO UTM algorithm can be used to generate paths for the fixed-wing aircraft or a combination of fixed wing and quadcopters.

 \begin{table}[h!]
\begin{center}
\caption{Path Length of five UAVs navingating the environment using RRT-VO UTM and RRT-APF UTM algorithm.}\label{tb:PathLength}
\begin{tabular}{|c||c||c|}
\hline
& RRT-VO ($m$) & RRT-APF ($m$)\\\hline
UAV 1 & 603.35 & 676.67\\
UAV 2 & 623.7 & --\\
UAV 3 & 534.77 & 632.73\\
UAV 4 & 469.23 & 616.39\\
UAV 5 & 696.16 & --\\\hline

\end{tabular}
\end{center}
\end{table}

 \begin{table}[h!]
\begin{center}
\caption{Path Length of seven UAVs navingating the environment using RRT-VO UTM and RRT-APF UTM algorithm.}\label{tb:PathLength7}
\begin{tabular}{|c||c||c|}
\hline
& RRT-VO ($m$) & RRT-APF ($m$)\\\hline
UAV 1 & 665.84 & 687.37\\
UAV 2 & 609.13 & 671.9091\\
UAV 3 & 604.9 & --\\
UAV 4 & 446.05 & --\\
UAV 5 & 608.17 & 724.66\\
UAV 6 & 612.51 & --\\
UAV 7 & 465.52 & 531.01\\\hline

\end{tabular}
\end{center}
\end{table}

\section{CONCLUSION}\label{CL}
A new UTM algorithm referred to as the RRT-VO UTM algorithm is developed by combining the Rapidly-Exploring Random Trees (RRT) and Velocity Obstacles (VO) algorithms. The tractability and effectiveness of this algorithm are demonstrated by generating safe and collision-free paths for multiple quadrotors. Furthermore, the comparison study indicates that the RRT-VO UTM algorithm outperforms the APF-based UTM algorithm in terms of safe navigation and lower path length. The proposed algorithm can also generate feasible paths for fixed-wing aircraft or a combination of fixed-wing and quadcopters. Future research will be focused on extending the RRT-VO UTM algorithm to generate feasible 3D paths for multiple UAVs.

\addtolength{\textheight}{-8cm}   


\section*{ACKNOWLEDGMENT}
Author's would like to acknowledge the funding  provided by ROCKWELL COLLINS (INDIA) ENTERPRISES PVT LTD, under the project "Online Planner for UTM" to carry out this research work.

\bibliographystyle{IEEEtran}
\bibliography{IEEEabrv, root}

\end{document}